\documentclass[10pt,journal,compsoc]{IEEEtran}
\ifCLASSOPTIONcompsoc
  \usepackage[nocompress]{cite}
  \usepackage{amsmath,amssymb,amsfonts}
	\usepackage{mathrsfs}
	\usepackage{algorithmic}
	\usepackage{algorithm}
	\usepackage{graphicx}
	\usepackage{textcomp}
	\usepackage{xcolor}
	\usepackage{booktabs}
	\usepackage{tabularx}
	\usepackage{array}
	\usepackage{multirow}	
	\DeclareMathOperator*{\argmin}{argmin}

	\def\BibTeX{{\rm B\kern-.05em{\sc i\kern-.025em b}\kern-.08em
    T\kern-.1667em\lower.7ex\hbox{E}\kern-.125emX}}
\else
  \usepackage{cite}
	\usepackage{amsmath,amssymb,amsfonts}
	\usepackage{mathrsfs}
	\usepackage{algorithmic}
	\usepackage{algorithm}
	\usepackage{graphicx}
	\usepackage{textcomp}
	\usepackage{xcolor}
	\usepackage{booktabs}
	\usepackage{tabularx}
	\usepackage{array}
	\usepackage{multirow}
	\DeclareMathOperator*{\argmin}{argmin}

	\def\BibTeX{{\rm B\kern-.05em{\sc i\kern-.025em b}\kern-.08em
    T\kern-.1667em\lower.7ex\hbox{E}\kern-.125emX}}
\fi
\ifCLASSINFOpdf
\else
\fi
\begin{document}
%
\title{AutoTS: Automatic Time Series Forecasting Model Design Based on Two-Stage Pruning}
%
%
%
%

\author{Chunnan~Wang,
        Xingyu~Chen,
        	Chengyue~Wu,
        and~Hongzhi~Wang$^\ast$,~\IEEEmembership{Member,~IEEE}
\IEEEcompsocitemizethanks{\IEEEcompsocthanksitem Chunnan Wang, Xingyu Chen, Chengyue Wu and Hongzhi Wang are with the Department of Computer Science and Technology, Harbin Institute of Technology, Harbin, China.\protect
\IEEEcompsocthanksitem Hongzhi Wang is the corresponding author. Chengyue Wu and Xingyu Chen contribute equally to this work.
\IEEEcompsocthanksitem Emails: \{WangChunnan, wangzh\}@hit.edu.cn, chenxylyf@126.com, 1190201314@stu.hit.edu.cn}
\thanks{Manuscript received None None, 2022; revised None None, 2022.}}

%
%

\markboth{Journal of \LaTeX\ Class Files,~Vol.~14, No.~8, August~2015}%
{Shell \MakeLowercase{\textit{et al.}}: Bare Demo of IEEEtran.cls for Computer Society Journals}
%



\IEEEtitleabstractindextext{%
\begin{abstract}
Automatic \underline{T}ime \underline{S}eries \underline{F}orecasting (TSF) model design which aims to help users to efficiently design suitable forecasting model for the given time series data scenarios, is a novel research topic to be urgently solved. In this paper, we propose AutoTS algorithm trying to utilize the existing design skills and design efficient search methods to effectively solve this problem. In AutoTS, we extract effective design experience from the existing TSF works. We allow the effective combination of design experience from different sources, so as to create an effective search space containing a variety of TSF models to support different TSF tasks. Considering the huge search space, in AutoTS, we propose a two-stage pruning strategy to reduce the search difficulty and improve the search efficiency. Specifically, at the beginning of the search phase, we apply the vertical pruning method to quickly optimize each module of the TSF model in turn. This method can quickly explore as many module options as possible, deeply extract the characteristics of  the search space components. In the middle of the search, we turn to apply the horizontal pruning method to filter out less effective options of each module according to the learned experience, and optimize multiple modules at the same time. This method breaks the limitation of sequential optimization and can explore more flexible model architectures. Two pruning methods complement each other and can guide AutoTS to explore the huge search space efficiently and reasonably. In addition, in AutoTS, we introduce the knowledge graph to reveal associations between module options. We make full use of these relational information to learn higher-level features of each module option, so as to further improve the search quality. Extensive experimental results show that AutoTS is well-suited for the TSF area. It is more efficient than the existing neural architecture search algorithms, and can quickly design powerful TSF model better than the manually designed ones.
\end{abstract}

}

\maketitle

\IEEEdisplaynontitleabstractindextext

%
\IEEEpeerreviewmaketitle

\IEEEraisesectionheading{\section{Introduction}\label{section：1}}

\IEEEPARstart{D}{ifferent} \underline{T}ime \underline{S}eries \underline{F}orecasting (TSF) models are often suitable for different data scenarios. How to quickly and automatically design the best TSF model for new data scenarios is a very important research topic in the field of time series, which is known as the automatic TSF model design. This research topic aims to help users break the technical restrictions, effectively reduce the cost of model development and efficiently deal with new time series data scenarios. It has high research value, but has not been studied yet. 

In this paper, we intend to fill this research gap. We propose the AutoTS algorithm, trying to make effective use of the existing design skills, and design an efficient automatic machine learning technology to realize the automatic design of the powerful TSF model. Our AutoTS algorithm consists of two parts: 
\begin{itemize}
\item \textbf{Part 1:} An effective search space with diversified TSF models.
\item \textbf{Part 2:} An efficient search strategy based on two-stage pruning and knowledge graph analysis.
\end{itemize}
The former one provides rich materials for AutoTS to effectively deal with a variety of time series tasks, and the latter one provides strong technical support for the efficient implementation of AutoTS.

In the search space part of AutoTS, we adopt the modular reorganization method for effective search space construction. We divide the TSF model architecture into multiple modules according to their functions. Then, we extract effective design options for each module from the existing works, and allow the effective combination of module options from different sources, so as to create an effective search space containing a variety of TSF models. This construction method can flexibly use the existing model design skills, and obtain diversified effective TSF model design schemes by integrating the advantages of different TSF works.

These schemes provide strong support for AutoTS to effectively deal with different TSF tasks. However, each module has a great amount of design options and the search space is huge. This brings great challenges to the optimal TSF model search. In order to find good architectures with relatively high probability under limited resources, a highly efficient search strategy, which can conduct more accurate sampling, i.e., exploring the subspace that contains potentially better architectures, is necessary. This is known as the space explosion issue in the \underline{N}eural \underline{A}rchitecture \underline{S}earch (NAS) area~\cite{DBLP:conf/cvpr/YuRS21}.

In recent years, researchers have proposed some progressive NAS algorithms, e.g., PNAS~\cite{DBLP:conf/eccv/LiuZNSHLFYHM18} and CNAS~\cite{DBLP:conf/icml/GuoCZZ0HT20}, to alleviate this issue. These NAS algorithms are designed for automatically and efficiently design the best CNN architecture. Their idea is to gradually enlarge the search space by adding deeper CNN or increasing the number of candidate options, and perform architecture search in a progressive manner. They can use the previously learned knowledge to make the sampling more accurate in the larger search space and thus improve the search quality.

These progressive algorithms perform well in the CNN-related NAS tasks, but are not suitable for our search space that is designed for the TSF model. Reasons are as follows. 
\begin{itemize}
\item \textbf{Failing to Add Deepth:} The position and function of the modules in the TSF model are fixed, and the model architecture cannot be deepened step by step. 
\item \textbf{Unsuited to Add Options:} Each module has a large number of different design options. In this case, gradually adding options can be very time-consuming, and the proper option adding sequence for each module is hard to be determined. 
\end{itemize}
The existing progressive search methods are obviously infeasible for our work. In AutoTS, we need to design more appropriate and efficient search methods to alleviate the space explosion issue caused by our TSF-based search space. 

We notice that in our designed TSF-based search space, there are many options in each module and different options may have very close relations. Considering these characteristics, in this paper, we propose a search method based on two-stage pruning and knowledge graph analysis. We try to (1) understand the characteristics of options in each module as quickly and fully as possible (2) and prune the search space reasonably and vigorously, so as to carry out efficient exploration.

Specifically, our proposed search method contains two stages, including vertical pruning and horizontal pruning. At the beginning of the search phase, we apply the vertical pruning method to sequentially optimize each module in the TSF model. This sequential optimization method can help us quickly explore as many module options as possible while searching the best model and deeply extract performance characteristics of the search space components.   

In the middle of the search, we turn to use the horizontal pruning method to effectively analyze the importance of each module option according to the learned characteristics, and filter out less important module options accordingly. At this stage, we break the restriction of sequential optimization, and optimize multiple modules at the same time. We can make full use of the learned performance evaluator to quickly search for a more flexible TSF model architecture from the lightweight search space. As we can see, two pruning methods are suitable for different search stages and complement each other. They can guide AutoTS to effectively explore the huge TSF model search space in an efficient and reasonable way.
 
In AutoTS, we also designed a knowledge graph for module options to help the search algorithm further understand the characteristics of each module option. Specifically, we notice that many module options in the search space have lots of overlaps in the hyperparameter setting and the architecture. We construct a knowledge graph to clarify these potential associations. We aim to make full use of these relational information to quickly grasp the composition characteristics of each module option, and thus obtain the higher-level representation of each option. With these higher-quality option representations, the search algorithm can better describe each TSF model, and thus make more effective decisions and improve the search quality.

Our contributions are summarized as follows:
\begin{itemize}
\item \textbf{Innovation:} We are the first to realize the automatic design of the TSF model. Our proposed AutoTS algorithm can greatly reduce the development cost of the TSF model, enabling non-expert users to easily access to high-performance TSF models.

\item \textbf{Efficiency:} Our proposed two-stage pruning and knowledge graph based search method can quickly and effectively deal with the huge model architecture search space, where there are many options for each module and options are closely related. 

\item \textbf{Effectiveness:} Extensive experimental results show that AutoTS can quickly design a TSF model, which is better than the existing manually designed TSF models. Moreover, compared with the existing NAS algorithms, AutoTS is more efficient and more suitable for the TSF area.
\end{itemize}

The remainder of this paper is organized into 5 sections. Section~\ref{section:2} introduces the existing TSF models and NAS algorithms. Section~\ref{section:3} gives the related concepts of the TSF task and definition of the automatic TSF model design problem. In Section~\ref{section:4}, we introduce the AutoTS algorithm in details. Section~\ref{section:5} evaluates the performance of AutoTS compared to existing TSF models and other NAS methods. Finally, we draw conclusions and present the future work in Section~\ref{section:6}. 

\section{Related Work}\label{section:2}

In this section, we introduce the existing TSF models and the NAS techniques, which are involved in this paper.

\subsection{Time Series Forecasting}\label{section:2.1}

\underline{T}ime \underline{S}eries \underline{F}orecasting (TSF) which aims to predict the trend of a time series has played an important role on many occasions of our daily life such as traffic, econometrics~\cite{DBLP:conf/aaai/YanXL18,DBLP:conf/iclr/LiYS018,DBLP:conf/iconip/SeoDVB18}. It has drawed much attention in recent years and many solutions have emerged to solve this problem. Among the existing solutions, deep learning methods~\cite{DBLP:conf/aaai/Lee20,DBLP:conf/ijcai/QinSCCJC17,DBLP:conf/nips/WuXDZ0H20,DBLP:conf/nips/VaswaniSPUJGKP17} perform the best and are widely used in the real applications. 

For example, LSTNet~\cite{DBLP:conf/sigir/LaiCYL18} uses the \underline{C}onvolution \underline{N}eural \underline{N}etwork (CNN) and the \underline{R}ecurrent \underline{N}eural \underline{N}etwork (RNN) to extract short-term local dependency patterns among variables and to discover long-term patterns for time series trends. \cite{DBLP:conf/iclr/OreshkinCCB20} proposed N-BEATS, a univariate model based on backward and forward residual links and a stack of fully-connected layers, to deal with the TSF tasks. N-BEATS is fast to train and is interpretable and applicable to a wide array of target domains. \cite{DBLP:conf/kdd/WuPL0CZ20} put forward the MTGNN which combines the \underline{G}raph \underline{N}eural \underline{N}etwork (GNN) and the \underline{T}emporal \underline{C}onvolutional \underline{N}etworks (TCN)~\cite{DBLP:conf/eccv/LeaVRH16}. MTGNN extracts the uni-directed relations among variables through the graph learning module. It can uses the mix-hop propagation layer and a dilated inception layer to get the further correlations within the time series. 

These existing works design various neural architectures to tackle  with the TSF tasks and have their own advantages. In this paper, we aim to flexibly use these valuable experience to support the automatic TSF model design. 

\begin{figure*}[tb]
	\centering
	\includegraphics[width=0.9\textwidth]{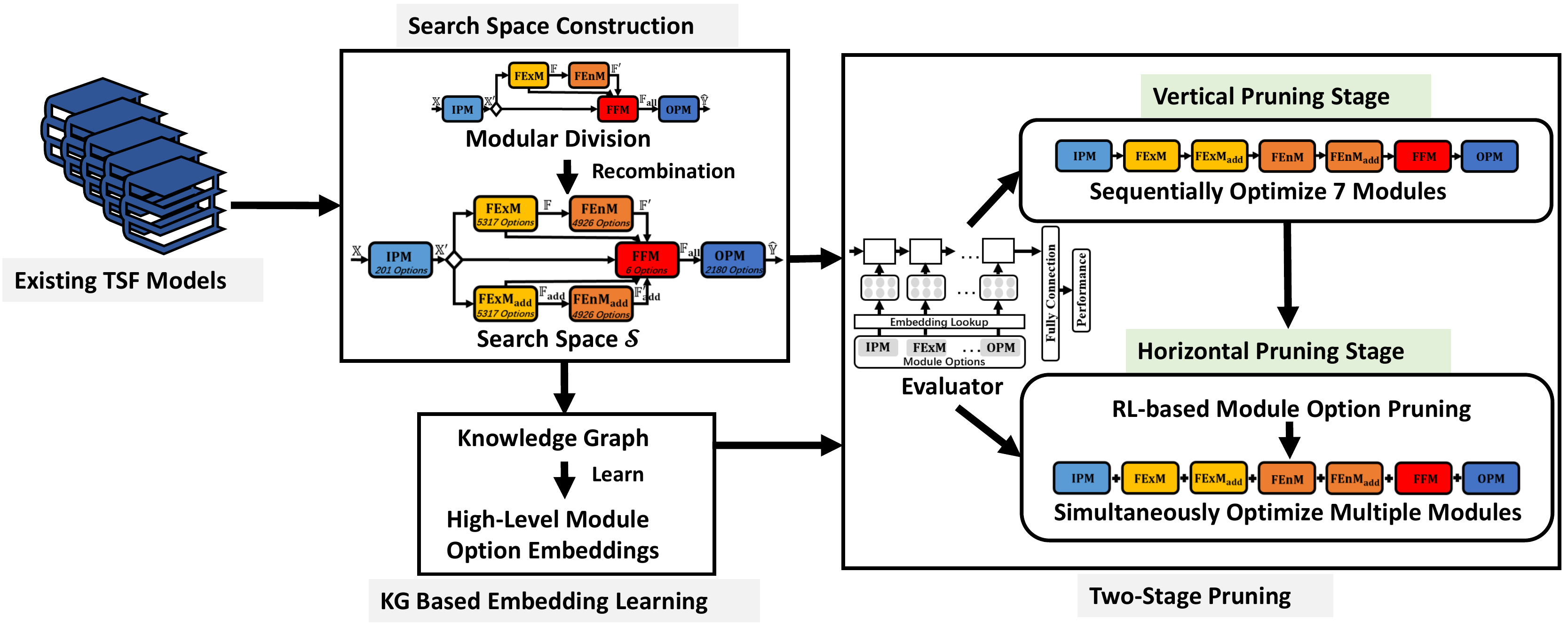}
	\caption{Overall framework of AutoTS. AutoTS learns composition and performance characteristics of module options through knowledge graph and vertical pruning respectively. Then, it efficiently explores good TSF models using the learned characteristics at the horizontal pruning stage.}
	\label{fig1}
\end{figure*}

\subsection{Neural Architecture Search}\label{section:2.2}

The \underline{N}eural \underline{A}rchitecture \underline{S}earch (NAS) which leans to automatically search for good neural architectures for the given dataset, is an important research topic in \underline{Auto}mated \underline{M}achine \underline{L}earning (AutoML). Many NAS algorithms have been proposed recently and they can be roughly classified into two categories in terms of search space.

The first category uses a fix search space throughout the search phase. This kind of NAS algorithms directly apply an effective search method, e.g., \underline{R}einforcement \underline{L}earning (RL) methods~\cite{DBLP:conf/ijcai/GaoYZ0H20, DBLP:conf/icml/BelloZVL17}, \underline{E}volutionary \underline{A}lgorithm (EA) based methods~\cite{DBLP:conf/cvpr/ChenMZXHMW19, DBLP:conf/aaai/RealAHL19} and gradient-based methods~\cite{DBLP:conf/iclr/LiuSY19, DBLP:conf/aistats/NoyNRZDFGZ20, DBLP:conf/cvpr/HosseiniYX21}, to explore the whole search space. The search space scale and the effectiveness of the applied search method have great influence on their performance. They are suitable for NAS problems with small-scale search space, and are widely used in the cell-based NAS problems which only search for the repeatable cell neural structures.

The second category applies a dynamic increasing search space. This kind of NAS algorithms (also known as the progressive NAS algorithms) gradually enlarge the search space by adding deeper neural architectures~\cite{DBLP:conf/eccv/LiuZNSHLFYHM18} or increasing the number of candidate options~\cite{DBLP:conf/icml/GuoCZZ0HT20}, and perform architecture search in a progressive manner. They can use previously learned knowledge to make the sampling more accurate in the larger search space and thus improve the search quality. They are well-suited for NAS problems with large-scale search space, but require the target neural architecture to be able to be deepened or the number of candidate options that are used for architecture construction is small. 

These exsiting NAS algoritgms have their own advantages and characteristics and are suitable for different kinds of NAS tasks. In this paper, we aim to find or design a suitable and efficient NAS method to deal with the automatic TSF model design problem, according to the characteristics of our designed TSF search space.

\section{Notations and Problem Definition}\label{section:3} 

In this section, we give the notations on TSF model and defines
the search target of AutoTS.

\textbf{Notations.} In this paper, we focus on the multivariate TSF tasks. Let $v_{t}\in\mathbb{R}^{N}$ be the value of a multivariate variable of dimension $N$ at time step $t$. The set of observed history observations of a multivariate variable over a time period $t_{obs}$ can be represented as  $\mathbb{X}=\{v_{1},v_{2},\cdots,v_{t_{obs}}\}\in\mathbb{R}^{N\times t_{obs}}$. A TSF model $\mathcal{M}$ can predict the upcoming observations of the multivariate variable over a future time horizon $t_{pred}$ , which is denoted by $\mathbb{Y}=\{v_{t_{obs}+1},v_{2},\cdots,v_{t_{pred}}\}$, according to $\mathbb{X}$. We use $\hat{\mathbb{Y}} = \mathcal{M}(\mathbb{X})$ to denote the prediction of model $\mathcal{M}$, and compare $\hat{\mathbb{Y}}$ with $\mathbb{Y}$ using \underline{R}oot \underline{R}elative \underline{S}quared \underline{E}rror (RRSE) or a loss function, so as to examine the effectiveness of the TSF model $\mathcal{M}$.

\textbf{Problem Definition.} Given a TSF search space $\mathcal{S}$ and a time series dataset $D$, the Automated TSF model design problem aims to find an optimal TSF model architecture $\mathcal{M}^{\ast}\in \mathcal{S}$ that minimizes the validation RRSE score on dataset $D$.
\begin{equation}
\begin{split}
&\mathcal{M}^{\ast}=\min_{\mathcal{M}\in \mathcal{S}}  RRSE_{val_{D}}(\mathbf{W}_{\mathcal{M}^{\ast}},\mathcal{M}) \\
&s.t.\ \mathbf{W}_{\mathcal{M}}^{\ast}= \mathop{\argmin}_{\mathbf{W}}  \mathcal{L}_{train_{D}} (\mathbf{W},\mathcal{M})
\end{split}
\end{equation}
where $\mathcal{L}_{train_{D}} (\mathbf{W},\mathcal{M})$ denotes the training loss of model $\mathcal{M}$ on $D$ under weights $\mathbf{W}$, and $train_{D}$ and $val_{D}$ denote the training set and the validation set of $D$ respectively.

\begin{table*}[htbp]
	\caption{Five types of modules that are used for TSF model construction. The third and the fourth columns give the options of each type of moudle.}
	\newcommand{\tabincell}[2]{\begin{tabular}{@{}#1@{}}#2\end{tabular}}
	\begin{center}
		\resizebox{0.95\textwidth}{!}{
\smallskip\begin{tabular}{m{2.6cm}|m{4.1cm}|m{4.5cm}|m{5.9cm}}
			\hline
			\textbf{Modules}&\textbf{Function Description}&\textbf{Existing solutions}&\textbf{Hyperparameters}\\
			\hline
			\multirow{2}{*}{\tabincell{l}{\textbf{Input Processing}\\ \textbf{Module ($\mathbf{IPM}$)}}}&\multirow{2}{*}{\tabincell{l}{Enhance the representation \\power of the input. \\ \\$\mathbb{X}'=\mathbf{IPM}(\mathbb{X})$\\ $\triangleright$ Number of options: 201}}&$\mathbf{IPM_{1}}$: Indentity Module ($\mathbf{T_{1}},\mathbf{T_{2}},\mathbf{T_{3}},\mathbf{T_{4}}$)&none \\
			\cline{3-4}
			&&$\mathbf{IPM_{2}}$: Graph Convolutional Network ($\mathbf{T_{5}}$)&gcn-true: \{True, False\}\newline node dim: \{10, 20, 30, 40\}\newline skip channels: \{8, 16, 32, 64, 96\}\newline residual channels: \{8, 16, 32, 64, 96\}\\
			\hline
			\multirow{5}{*}{\tabincell{l}{\textbf{Feature Extraction}\\ \textbf{Module ($\mathbf{FExM}$)}}}& \multirow{5}{*}{\tabincell{l}{Extract the features of the \\historical time series.\\ \\ $\mathbb{F}=\mathbf{FExM}(\mathbb{X}')$\\ $\mathbb{F}_{add}=\mathbf{FExM_{add}}(\mathbb{X}')$\\ $\triangleright$ Number of options: 5317}}&$\mathbf{FExM_{1}}$: Identity Module ($\mathbf{T_{1}},\mathbf{T_{3}}$)&none
			\\
			\cline{3-4}
			&&$\mathbf{FExM_{2}}$: Temporal Pattern Attention based LSTM ($\mathbf{T_{2}}$)&number of layers: \{1, 2, 3, 4, 5\}
			\newline number of units: \{128, 256, 384, 512, 768, 896\}\\
			\cline{3-4}
			&&$\mathbf{FExM_{3}}$: Generic Trend and Seasonality Block $(\mathbf{T_{4}})$& thetas dim: \{(1,1), (2,2), (3,3), (1,2), (1,3), (2,3)\} 
			\newline hidden layer units: \{128, 256, 384, 512, 768, 896\} 
			\newline stack type: \{'trend', 'generic', 'seasonality'\} 
			\newline stack id: \{0, 1\}
			\newline number of blocks : \{1, 2, 3, 4, 5, 6, 7, 8, 9, 10\} 
			\\
			\cline{3-4}
			&&$\mathbf{FExM_{4}}$: Temporal Convolutional Layer ($\mathbf{T_{5}}$)&conv channels: \{8, 16, 32, 64, 96\} 
			\newline gcn depth: \{1, 2, 3, 4, 5\} 
			\newline number of layers: \{1, 2, 3, 4, 5\}
			\newline skip channels: \{8, 16, 32, 64, 96\} 
			\newline residual channels: \{8, 16, 32, 64, 96\}
			\\
			\cline{3-4}
			&&$\mathbf{FExM_{5}}$: none &none\\
			\hline
			
			\multirow{3}{*}{\tabincell{l}{\textbf{Feature Enhancement}\\ \textbf{Module ($\mathbf{FEnM}$)}}}&\multirow{3}{*}{\tabincell{l}{Enhance representation of the\\ extracted features.\\ \\ $\mathbb{F}'=\mathbf{FEnM}(\mathbb{F})$\\ $\mathbb{F}_{add}'=\mathbf{FEnM_{add}}(\mathbb{F}_{add})$\\ $\triangleright$ Number of options: 4926}}&$\mathbf{FEnM_{1}}$: Identity Model ($\mathbf{T_{1}},\mathbf{T_{2}},\mathbf{T_{4}}$)
			& none
			
			\\
			\cline{3-4}
			&&$\mathbf{FEnM_{2}}$: Residual Block ($\mathbf{T_{3}}$)&kernel size: \{2, 3, 4, 5, 6, 7\}
			\newline number of filters: \{2, 3, 4, 5, 6, 7\}
			\newline dilation base: \{1, 2, 3, 4, 5\}
			\newline weight norm: \{True, False\}
			\newline dropout: \{0, 0.05, 0.1, 0.15, 0.2\}
			\\
			\cline{3-4}
			&&$\mathbf{FEnM_{3}}$: Temporal and Graph Convolution ($\mathbf{T_{5}}$)&conv channels: \{8, 16, 32, 64, 96\} 
			\newline gcn depth: \{1, 2, 3, 4, 5\} 
			\newline number of layers: \{1, 2, 3, 4, 5\}
			\newline skip channels: \{8, 16, 32, 64, 96\} 
			\newline residual channels: \{8, 16, 32, 64, 96\}
			\\
			\hline
			\multirow{2}{*}{\tabincell{l}{\textbf{Feature Fusion}\\ \textbf{Module ($\mathbf{FFM}$)}}}&\multirow{2}{*}{\tabincell{l}{Combine all features.\\ $\mathbb{F}_{all}=\mathbf{FFM}(\mathbb{X}, \mathbb{F},\mathbb{F}_{add},\mathbb{F}',\mathbb{F}_{add}')$\\$\triangleright$ Number of options: 6}}&$\mathbf{FFM_{1}}$: Identity Module\newline($\mathbf{T_{1}},\mathbf{T_{2}},\mathbf{T_{3}},\mathbf{T_{4}}$)&none
			\\
			\cline{3-4}

			&&$\mathbf{FFM_{2}}$: Skip Connection Module ($\mathbf{T_{5}}$) & skip channels: \{8, 16, 32, 64, 96\}\\
			\hline
			\multirow{5}{*}{\tabincell{l}{\textbf{Output Processing}\\ \textbf{Module ($\mathbf{OPM}$)}}}& \multirow{5}{*}{\tabincell{l}{Transform the combined \\features to the expected\\ predictions.\\ \\ $\hat{\mathbb{Y}}=\mathbf{OPM}(\mathbb{F}_{all})$\\ $\triangleright$ Number of options: 2180}}&$\mathbf{OPM_{1}}$: Dense ForecastNet ($\mathbf{T_{1}}$) &hidden dim: \{24, 36, 48, 60, 72, 84, 96, 128\} \\
			\cline{3-4}
			&&$\mathbf{OPM_{2}}$: Fully-Connected Layer ($\mathbf{T_{2}}$)&
			opt num units: \{128, 256, 384, 512, 768, 896\}
			\\
			\cline{3-4}
			&&$\mathbf{OPM_{3}}$: Identity Module ($\mathbf{T_{3}}$)&none\\
			\cline{3-4}
			&&$\mathbf{OPM_{4}}$: Generic Trend and Seasonality Block ($\mathbf{T_{4}}$)&thetas dim: \{(1,1), (2,2), (3,3), (1,2), (1,3), (2,3)\} 
			\newline hidden layer units: \{128, 256, 384, 512, 768, 896\} 
			\newline stack type: \{'trend', 'generic', 'seasonality'\} 
			\newline stack id: \{0, 1\}
			\newline number of blocks : \{1, 2, 3, 4, 5, 6, 7, 8, 9, 10\} 
			\\
			\cline{3-4}
			&&$\mathbf{OPM_{5}}$: Convolutional Layer ($\mathbf{T_{5}}$)&end channels: \{8, 16, 32, 64, 96\} 
			\\
			\hline
		\end{tabular}
		}
		\label{table1}
	\end{center}
\end{table*}

\section{AutoTS Algorithm}\label{section:4}

In this paper, we present AutoTS to achieve the automatic design of the TSF model. First, we utilize the experience provided by the existing TSF models to construct a huge search space with diversified TSF model architectures (Section~\ref{section:4.1}). Then, we present the two-stage pruning method to efficiently explore the huge search space and effectively search for the powerful TSF model for the given dataset (Section~\ref{section:4.2}). Finally, we introduce the knowledge graph embedding method that is used for obtaining higher-level option embedding and further improving the search performance in AutoTS (Section~\ref{section:4.3}). Fig.~\ref{fig1} gives the overall framework of AutoTS.

\subsection{Search Space}\label{section:4.1}

In AutoTS, we unpack the TSF model architecture into five parts by learning from the existing TSF models: 
\begin{itemize}
\item Input Processing Module ($\mathbf{IPM}$), which enhances the representation power of the input.\\ $\mathbb{X}'=\mathbf{IPM}(\mathbb{X})$
\item Feature Extraction Module ($\mathbf{FExM}$), which captures features of the historical time series. \\ $\mathbb{F}=\mathbf{FExM}(\mathbb{X}')$
\item Feature Enhancement Module ($\mathbf{FEnM}$), which  enhances representation of the extracted features. \\ $\mathbb{F}'=\mathbf{FEnM}(\mathbb{F})$
\item Feature Fusion Module ($\mathbf{FFM}$), which combines all features of the historical time series. \\ $\mathbb{F}_{all}=\mathbf{FFM}(\mathbb{F}, \mathbb{F}')$
\item Output Processing Module ($\mathbf{OPM}$), which transforms the combined features into the expected prediction. \\ $\hat{\mathbb{Y}}=\mathbf{OPM}(\mathbb{F}_{all})$
\end{itemize}
We extract effective solutions of each moudle from 5 state-of-the-art TSF models, including \textbf{ForecastNet} ($\mathbf{T_{1}}$)~\cite{DBLP:conf/iconip/DabrowskiZR20}, \textbf{Temporal Pattern Attention} ($\mathbf{T_{2}}$)~\cite{DBLP:journals/ml/ShihSL19}, \textbf{Temporal Convolutional Networks} ($\mathbf{T_{3}}$)~\cite{DBLP:conf/eccv/LeaVRH16}, \textbf{N-BEATS} ($\mathbf{T_{4}}$)~\cite{DBLP:conf/iclr/OreshkinCCB20} and \textbf{MTGNN}($\mathbf{T_{5}}$)~\cite{DBLP:conf/kdd/WuPL0CZ20}. Table~\ref{table1} summarizes these solutions and their hyperparameters.

\begin{figure}[tb]
	\centering
	\includegraphics[width=0.5\textwidth]{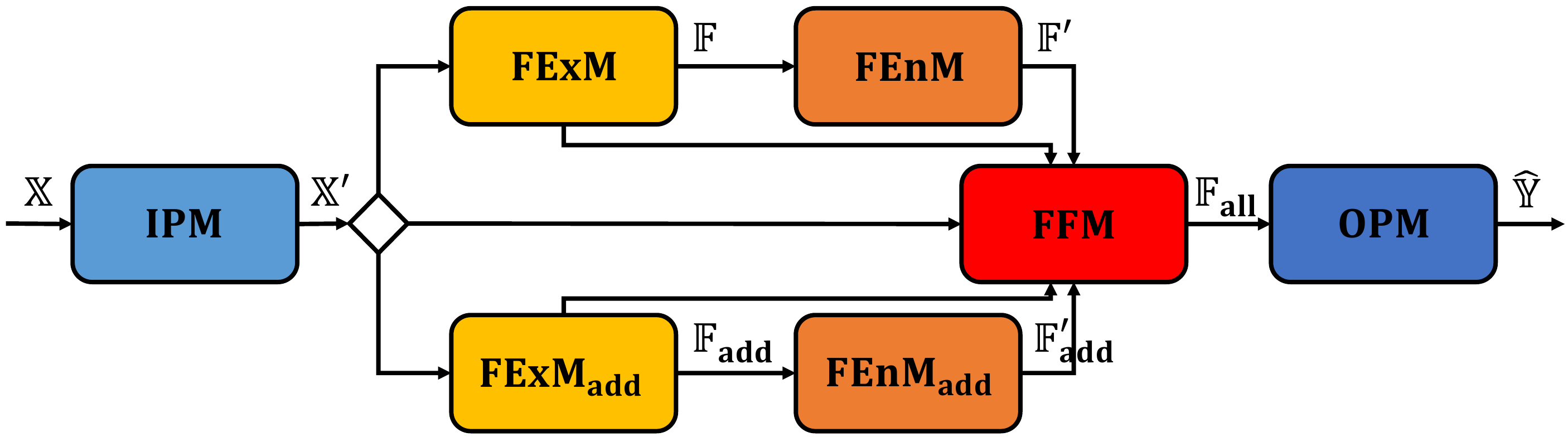}
	\caption{The structure of the TSF model in our search space $\mathcal{S}$. Each TSF model in $\mathcal{S}$ is composed of 7 modules, whose options are given in the third and the fourth columns of Table~\ref{table1}.}
	\label{fig2}
\end{figure}

In AutoTS, we utilize the experience information in Table~\ref{table1} to construct an effective TSF search space. Specifically, apart from 5 models defined above which outline the overall architecture of the existing TSF model, we add two modules: $\mathbf{FExM_{add}}$ and $\mathbf{FEnM_{add}}$ to TSF models in our search space $\mathcal{S}$, applying dual-channel feature extraction-enhancement structure to acquire more stronger fitting ability. We use these 7 modules to construct a complete TSF model. Fig.~\ref{fig2} illustrates the architecture of the TSF models in our search space $\mathcal{S}$. 

Considering that hyperparameters also affect the final model structure and performance, we take the module solution with different hyperparameter settings as different module options in AutoTS (module options are shown in the third and the fourth column of Table~\ref{table1}). We allow each of the module to be set to the corresponding option, and thus obtain a search space $\mathcal{S}$ with diversified TSF models. Our proposed search space considers both the hyperparameter setting diversity and the source diversity. It can make flexible use of the existing design experience to create many novel and more valuable TSF model architectures for different TSF tasks. 

\textbf{Scale of the Search Space.} The second column of Table~\ref{table1} gives the number of options for each module. We can see that most modules in the TSF model have a large number of options to choose from, e.g., $\mathbf{FExM}$ and $\mathbf{IPM}$ have 5,317 and 2,180 options respectively. Multiply the option numbers of 7 modules in Table~\ref{table1}, then we get the scale of our search space $\mathcal{S}$: $|\mathcal{S}|\approx1.8\times 10^{21}$. 

\textbf{Characteristics of the Search Space.} Our constructed TSF-oriented search sapce $\mathcal{S}$ have the following two characteristics:
\begin{itemize}
\item \textbf{C1: Large Module Search Sapce.} Most modules in the TSF model have a large number of options.
\item \textbf{C2: Multiple Modules.} Each TSF model in the search sapce consists of multiple moudles. 
\item \textbf{C3: Huge Model Search Space.} The search space contains huge amount of TSF models.
\end{itemize}
\textbf{C3} makes $\mathcal{S}$ not suitable for NAS methods with a fix search space and \textbf{C1} makes $\mathcal{S}$ fail to cooperate with the exitsing progressive NAS methods, as discussed in Section~\ref{section:2.2}. Obviously, $\mathcal{S}$ is not suitable for the existing NAS methods. We need to design an efficient and suitable search method, and thus quickly find high-quality TSF models from $\mathcal{S}$. 

\begin{figure*}[tb]
	\centering
	\includegraphics[width=0.75\textwidth]{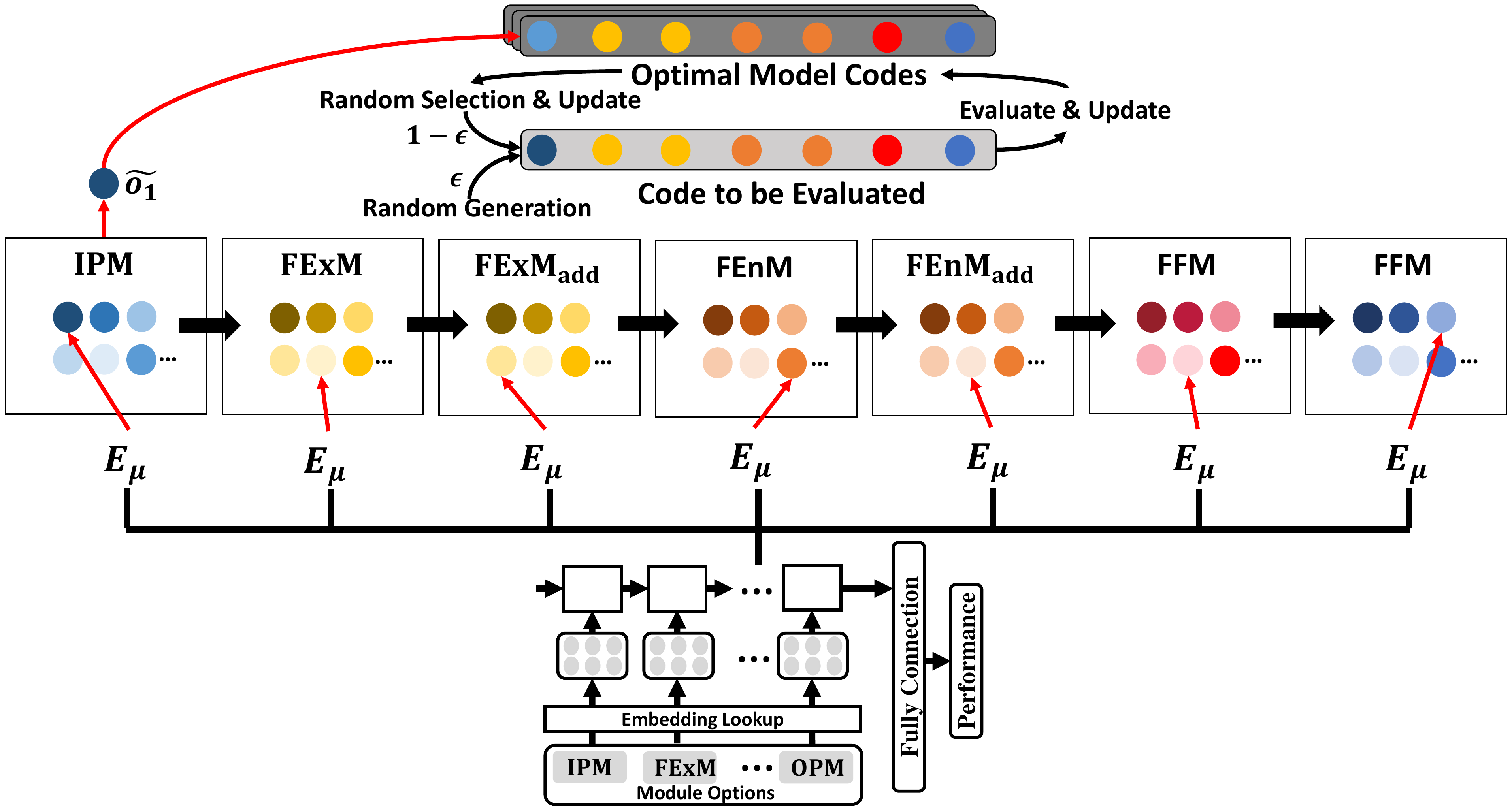}
	\caption{Illustration of the vertical pruning method. This method aims to search for better TSF models and fully explores moudle options by sequentially optimizing different modules of the well-performed model code. The circles with different colors corresponds to different module options.}
	\label{fig3}
\end{figure*}

\subsection{Two-Stage Pruning}\label{section:4.2}

Considering the characteristics of our TSF search space, in AutoTS, we design the two-stage pruning search method to improve the search efficiency. We aim to increase the possibility of finding high-performance TSF models from the huge search sapce $\mathcal{S}$ by:
\begin{itemize}
\item \textbf{(1) Feature Understanding.} Understanding the characteristics of options in each module as quickly and fully as possible;
\item \textbf{(2) Space Pruning.} Reduce the search space scale reasonably and vigorously.
\end{itemize}

\textbf{Design Idea.} Specifically, considering \textbf{C1} and \textbf{C2} of $\mathcal{S}$, in AutoTS, we can reduce the search space scale either by reducing the number of optimized modules or by reducing the number of options for each module. We call the former method the vertical pruning and call the latter one the horizontal pruning. 

At the vertical pruning stage, we can optimize the specific module of the TSF model in turn and thus changing the search space scale to $\sum_{i=1}^{7}|\mathcal{O}_{i}|$ (we use $\mathcal{O}_{i}$ to denote the option set of the $i^{th}$ moudle of the TSF model for simplicity). This pruning method can enable us to efficiently explore module options in the search space $\mathcal{S}$ and learn their high-order characteristics. But it can not optimize different modules at the same time, which is  inflexible.

At the vertical pruning stage, we can greatly cut the search space of each module and thus change the search space scale to $\prod_{i=1}^{7}|\tilde{\mathcal{O}_{i}}|(|\tilde{\mathcal{O}_{i}}|\ll|\mathcal{O}_{i}|)$. This pruning method can make up for the deficiency of the vertical pruning, enabling us to explore more flexible module option combinations. But, it needs lots of experience for reasonable option pruning.

We can notice that the two pruning methods complement each other. The vertical pruning does not rely on lots of experience and is more suitable for the early stage of search phase. It can learn lots of experience during the search phase and provide support for subsequent operations. The horizontal pruning is suitable for the later stage of search phase. It can effectively prune less important module options according to the learned experience, and thus quickly explore more diverse TSF models. Based on this idea, we present a two-stage pruning method implementing two pruning strategies in turn, so as to guide AutoTS to explore the huge search space efficiently and reasonably. Details are as follows.

\subsubsection{Vertical Pruning Stage}\label{section:4.2.1}

\textbf{Goal and Symbols.} At the vertical pruning stage, we aim to sequentially optimize 7 modules of the TSF model for each iteration. We denote the set of the searched Top $K$ model codes as $\mathbf{BestMs}$. In addition, we build an evaluator $\mathbf{E_{\mu}}$, which is composed of a LSTM layer and a fully-connected layer, to predict the performance score, i.e., validation RRSE score $RRSE_{val_{D}}$, of a certain model code. 

\textbf{Workflow Details.} Each time we randomly select a well-performed model code $\mathbf{BestM}=(o_{1},\ldots, o_7)$ from $\mathbf{BestMs}$, where $o_{i}\in \mathcal{O}_{i}$ is an option for the $i^{th}$ module of $\mathbf{BestM}$, and optimize one module of $\mathbf{BestM}$ under the guidance of the learned evaluator $\mathbf{E_{\mu}}$. Taking the $i^{th}$ module as an example. We use the evaluator $\mathbf{E_{\mu}}$ to find the best option $\tilde{o_{i}}$ to replace the corresponding module options in $\mathbf{BestOs}$, so as to construct a better TSF model code, which is denoted as  $\widetilde{NewM}$.
\begin{equation}
\begin{split}
\tilde{o_{i}}= &\mathop{\argmin}_{o\in\mathcal{O}_{i}}\mathbf{E_{\mu}}\big(\mathbf{Emb_{\theta}}(\mathbf{BestM}_{1\sim i-1}, o, \mathbf{BestM}_{i+1\sim 7})\big)\\
&\widetilde{NewM}=(\mathbf{BestM}_{1\sim i-1}, \tilde{o_{i}}, \mathbf{BestM}_{i+1\sim 7})
\end{split}
\end{equation}
where $\mathbf{Emb_{\theta}}(*)$ represents the embeddings of the options in the model code $*$, whose learning method will be introduced in Section~\ref{section:4.3}. The evaluator $E_{\mu}$ can extract the high-level feature of the given model code and predict its performance accordingly. 

After obtaining $\widetilde{NewM}$, we calculate its validation RRSE score which is denoted by $\widetilde{NewP}$ and add it to the historical evaluation data $\mathcal{D}$:
\begin{equation}
\mathcal{D}=\mathcal{D}\cup\{(\widetilde{NewM},\widetilde{NewP})\}
 \end{equation} 
We then optimize the parameters of $\mathbf{E_{\mu}}$ and $\mathbf{Emb_{\theta}}$ and update $\mathbf{BestMs}$ using $\mathcal{D}$: 
\begin{equation}
\nabla_{\mathcal{\theta},\mathcal{\mu}} \sum_{(M,P)\in \mathcal{D}} \Big(\mathbf{E_{\mu}}\big(\mathbf{Emb_{\theta}}(M)\big)-P\Big)^{2}
\label{equ4}
\end{equation} 
\begin{equation}
\mathbf{BestMs}= TopK\ model\ codes\ in\ \mathcal{D}
\label{equ5}
\end{equation} 
 
At the vertical pruning stage of AutoTS, we mainly alternatively optimize each module of the TSF model according to the above steps. In addition, we also introduce the $\epsilon$-greedy strategy to randomly generated and evaluate model codes with a certain probability, so as to avoid falling into local optimum. Algorithm~\ref{alg1} gives the pseudo code of the vertical pruning method and Fig.~\ref{fig3} illustrates its process. 

\textbf{Advantage Analysis.} The vertical pruning method searches for a better TSF model and fully explores moudle options by optimizing the options of each module of the well-performed model code in turn. It can quickly learn the effective representation $\mathbf{Emb_{\theta}}$ of each module option, and obtain high-quality performance evaluator $\mathbf{E_{\mu}}$. These experience information is pretty helpful for horizontal pruning and can provide favorable support for effective module option pruning and more flexible TSF model search.

\begin{algorithm}
	\caption{Vertical Pruning Method}
	\begin{algorithmic}[1] 
		\REQUIRE Search space $\mathcal{S}=\mathcal{O}_{1}\wedge\mathcal{O}_{2}\wedge\cdots\wedge\mathcal{O}_{7}$, 
		Search time $\mathcal{T}$, Number of top models $K$, Random possibility $\epsilon$, Embeddings of options $\mathbf{Emb}_{\theta}\in \mathbb{R}^{N\times \sum_{i=1}^{7} |\mathcal{O}_{i}|}$

	    \STATE $i\gets 0$
	    \STATE $\mathbf{BestMs}\gets$ Initilize $\mathbf{BestMs}$ by randomly select $K$ model codes from $\mathcal{S}$ and evaluate them
	    \WHILE{search time $< \frac{\mathcal{T}}{2}$}  
	    		\STATE $i\gets mode(i,7)+1$
	    		\IF{random number $<\epsilon$}
		    		\STATE $\widetilde{NewM}\gets$ Randomly select a code from $\mathcal{S}$
		    		\ELSE
		    		\STATE $\mathbf{BestM}\gets$ Randomly select a code from $\mathbf{BestMs}$
		    		\STATE $\tilde{o_{i}}\gets \mathop{\argmin}\limits_{o\in\mathcal{O}_{i}}\mathbf{E_{\mu}}\big(\mathbf{Emb_{\theta}}(\mathbf{BestM}_{1\sim i-1}, o, \mathbf{BestM}_{i+1\sim 7})\big)$
		    		\STATE $\widetilde{NewM}\gets (\mathbf{BestM}_{1\sim i-1}, \tilde{o_{i}}, \mathbf{BestM}_{i+1\sim 7})$
	    		\ENDIF
	    		\STATE $\widetilde{NewP}\gets$  Performance of $\widetilde{NewM}$
	    		\STATE $\mathcal{D}\gets \mathcal{D}\cup\{(\widetilde{NewM},\widetilde{NewP})\}$
	    		\STATE $\mathbf{BestMs}\gets$ Top-$K$ model codes in $\mathcal{D}$
	    		\STATE Optimize $\mathbf{E_{\mu}}$ and fine tune $\mathbf{Emb_{\theta}}$ using $\mathcal{D}$ according to Equation~\ref{equ4}
	    	\ENDWHILE
    	\RETURN $\mathbf{BestMs}$, $\mathbf{Emb}_{\theta}$, $\mathcal{D}$
\end{algorithmic}  
\label{alg1}
\end{algorithm}

\subsubsection{Horizontal Pruning Stage}

\begin{figure*}[tb]
	\centering
	\includegraphics[width=0.75\textwidth]{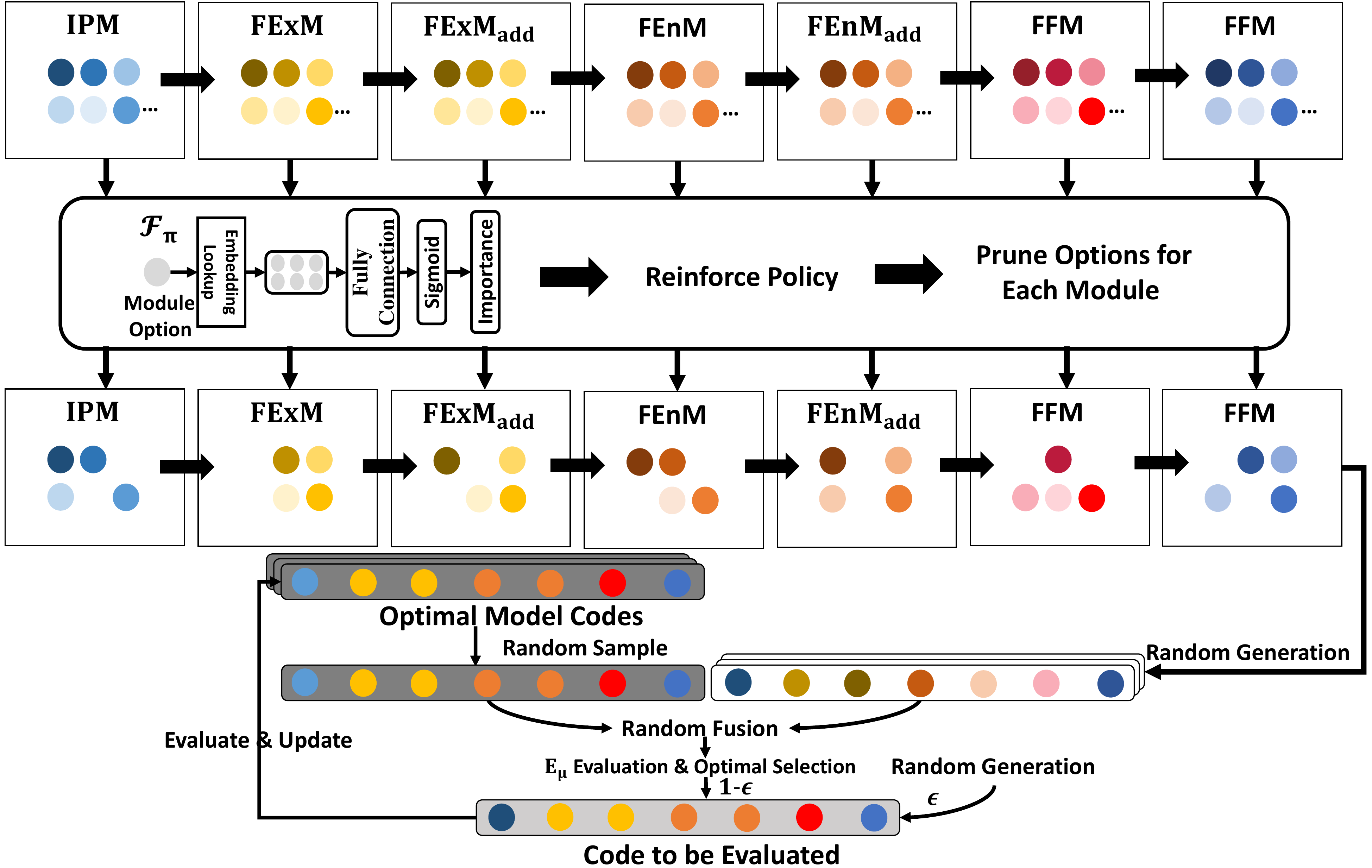}
	\caption{Illustration of the horizontal pruning method. This method aims to greatly cut the search space of each module and thus efficiently explore more flexible model code. The circles with different colors corresponds to different module options.
	}
	\label{fig4}
\end{figure*}

\textbf{Goal and Symbols.} At the horizontal pruning stage, we aim to efficiently and simultaneously optimize multiple modules of the TSF model. We build an importance predictor $\mathbf{F_{\pi}}$, which is composed of a fully connected layer, to predict the importance reasonably according to the high-order representation of each module option. 

\textbf{Workflow Details.} We introduce reinforce policy~\cite{p12, p22} to optimize $\mathbf{F_{\pi}}$ by maximizing the reward expectation, i.e., by improving the probability of selecting high-quality TSF models:
\begin{equation}
\small
\nabla_{\pi}\sum_{\substack{(M,P)\in \mathcal{D}\\M=(o_1,...,o_7)}} (P_{max}-P)\times log\Big(\frac{1}{7} \sum_{i=1}^{7}\frac{\mathbf{F_{\pi}}\big(\mathbf{Emb_{\theta}}(o_{i})\big)}{\sum\limits_{o'\in\mathcal{O}_{i}}\mathbf{F_{\pi}}\big(\mathbf{Emb_{\theta}}(o')\big)}\Big)
\label{equ6}
\end{equation}
where $(P_{max}-P)$ is the reward of model code $M$, $P_{max}$ is the upper limits of the validation RRSE score, and the formula in log denotes the possibility of obtaining this reward. After getting the well-trained importance predictor $\mathbf{F_{\pi}}$, we filter out some excellent options for each module according to the calculated option importance value to construct a lightweight search space: $\tilde{\mathcal{S}}=\tilde{\mathcal{O}_{1}}\wedge\cdots\wedge\tilde{\mathcal{O}_{7}}$, where $\tilde{\mathcal{O}_{i}}\subseteq\mathcal{O}_{i}$ and $|\tilde{\mathcal{O}_{i}}|<<|\mathcal{O}_{i}|$.

As for the TSF model optimization, at the horizontal pruning stage, we utilize the learned performance evaluator $\mathbf{E_{\mu}}$ and the well-performed model codes $BestMs$ to efficiently search for a better TSF model, which is denoted as $\widetilde{NewM}$, from the prunned search space $\tilde{\mathcal{S}}$. Specifically, for each iteration, we randomly select a well-performed model code $\mathbf{BestM}=(o_{1},\ldots, o_7)$ from $\mathbf{BestMs}$. We take $\mathbf{BestM}$ as a base model code, randomly adjust it using $\tilde{\mathcal{S}}$, and utilize the evaluator $\mathbf{E_{\mu}}$ to find the best adjustment method. 
\begin{equation}
\widetilde{NewM}= \mathop{\argmin}_{M\in\tilde{\mathcal{S}}} \mathbf{E_{\mu}}\Big(\mathbf{Emb_{\theta}}\big(\mathbf{RFusion}(\mathbf{BestM},M)\big)\Big)
\end{equation}
where $\mathbf{RFusion}(*,\star)$ is a random fusion operation which randomly selects module options from $\star$ to adjust $*$ for generating a new model code. As we can see, with the help of the pruned search space and the well-trained evaluator and embeddings, the horizontal pruning method can more flexibly and effectively optimize well-performed model codes.

\begin{algorithm}
	\caption{Horizontal Pruning Method}
	\begin{algorithmic}[1] 
		\REQUIRE Search space $\mathcal{S}=\mathcal{O}_{1}\wedge\mathcal{O}_{2}\wedge\cdots\wedge\mathcal{O}_{7}$, 
		Search time $\mathcal{T}$, Number of top models $K$, Random possibility $\epsilon$, Prune ratio $\gamma$, $\mathbf{BestMs}$, $\mathbf{Emb}_{\theta}$, $\mathcal{D}$

	    \STATE Optimize $\mathbf{F_{\pi}}$ using $\mathcal{D}$ according to Equation~\ref{equ6}
	    \STATE $\tilde{\mathcal{O}_{i}}\gets$ Top $\gamma\times |\mathcal{O}_{i}|$ options with the highest importance scores $\mathbf{F_{\pi}}\big(\mathbf{Emb}_{\theta}(o)\big)$ $(o\in \mathcal{O}_{i}, i=1,...,7)$
	    \STATE $\tilde{\mathcal{S}}\gets \tilde{\mathcal{O}_{1}}\wedge\cdots\wedge\tilde{\mathcal{O}_{7}}$
	    \WHILE{search time $< \frac{\mathcal{T}}{2}$}  
	    		\IF{random number $<\epsilon$}
		    		\STATE $\widetilde{NewM}\gets$ Randomly select a code from $\mathcal{S}$
		    		\ELSE
		    		\STATE $\mathbf{BestM}\gets$ Randomly select a code from $\mathbf{BestMs}$
		    		\STATE $\widetilde{NewM}\gets \mathop{\argmin}\limits_{M\in\tilde{\mathcal{S}}} \mathbf{E_{\mu}}\Big(\mathbf{Emb_{\theta}}\big(\mathbf{RFusion}(\mathbf{BestM},M)\big)\Big)$
	    		\ENDIF
	    		\STATE $\widetilde{NewP}\gets$  Performance of $\widetilde{NewM}$
	    		\STATE Update $\mathcal{D}$ and $\mathbf{BestMs}$, adjust $\mathbf{E_{\mu}}$ and $\mathbf{Emb_{\theta}}$ according to Equation~\ref{equ4}
	    		\STATE  Optimize $\mathbf{F_{\pi}}$ using $\mathcal{D}$ according to Equation~\ref{equ8}
	    		\STATE  $\tilde{\mathcal{O}_{i}}\gets$ Top $\gamma\times |\mathcal{O}_{i}|$ options with the highest importance scores $\mathbf{F_{\pi}}\big(\mathbf{Emb}_{\theta}(o)\big)$ $(o\in \mathcal{O}_{i}, i=1,...,7)$
	    		\STATE $\tilde{\mathcal{S}}\gets \tilde{\mathcal{O}_{1}}\wedge\cdots\wedge\tilde{\mathcal{O}_{7}}$
	    	\ENDWHILE
    	\RETURN $\mathbf{BestMs}$
\end{algorithmic}  
\label{alg2}
\end{algorithm}

After getting $\widetilde{NewM}$, we calculate its validation RRSE score which is denoted by $\widetilde{NewP}$ and add it to the historical evaluation data $\mathcal{D}$. With the new evaluation data, we adjust $\mathbf{E_{\mu}}$, $\mathbf{Emb_{\theta}}$ and $\mathbf{BestMs}$ according to Equation~\ref{equ4} and Equation~\ref{equ5}, and optimize the parameters of the predictor $\mathbf{F_{\pi}}$ as follows:
\begin{equation}
\small
\nabla_{\pi}\sum_{\substack{(M,P)\in \mathcal{D}_{batch}\\M=(o_1,...,o_7)}} (P_{max}-P)\times log\Big(\frac{1}{7} \sum_{i=1}^{7}\frac{\mathbf{F_{\pi}}\big(\mathbf{Emb_{\theta}}(o_{i})\big)}{\sum\limits_{o'\in\tilde{\mathcal{O}_{i}}}\mathbf{F_{\pi}}\big(\mathbf{Emb_{\theta}}(o')\big)}\Big)
\label{equ8}
\end{equation}
where $\mathcal{D}_{batch}$ denotes a batch data of $\mathcal{D}$ which concludes $(\widetilde{NewM}, \widetilde{NewP})$. With the optimized $\mathbf{F_{\pi}}$, the pruned search space $\tilde{\mathcal{S}}=\tilde{\mathcal{O}_{1}}\wedge\cdots\wedge\tilde{\mathcal{O}_{7}}$ can be updated. We repeat the above optimization steps to explore better model codes using the new search space until the search is complete. Algorithm~\ref{alg2} and Fig.~\ref{fig4} give the pseudo code and illustration of the horizontal pruning method. Note that, same as the vertical pruing method, we introduce $\epsilon$-greedy strategy to avoid falling into local optimum.

\textbf{Advantage Analysis.} The horizontal pruning method continuously optimizes the search space pruning scheme, and use the learned performance evaluator $\mathbf{E_{\mu}}$ to quickly explore the pruned search space $\tilde{\mathcal{S}}$ to obtain better TSF models. With the help of small-scale search space and well-trained evaluator, it can efficiently and flexibly explore more diverse TSF model architectures and thus further improve the search quality.

\subsection{Embedded Learning Based on Knowledge Graph}\label{section:4.3}

We notice that the representations of the module options $\mathbf{Emb_{\theta}}$ play an important role in the two-stage pruning of AutoTS. $\mathbf{F_{\pi}}$ uses $\mathbf{Emb_{\theta}}$ to analyze the importance of each module option and $\mathbf{E_{\mu}}$ utilizes $\mathbf{Emb_{\theta}}$ to analyze the model characteristics and performance. The quality of $\mathbf{Emb_{\theta}}$ greatly affects the effectiveness of $\mathbf{F_{\pi}}$ and $\mathbf{E_{\mu}}$. Therefore, the high-level module option embeddings $\mathbf{Emb_{\theta}}$ is necessary for our algorithm.

In the two-stage pruning phase, we only obtain the $\mathbf{Emb_{\theta}}$ by fitting the mapping relationship between the TSF model code and its performance (Equation~\ref{equ4}). This method can learn the performance characteristics of each option, but is not desirable enough. 

\textbf{Design Idea.} We find that the options of each module have close correlation, and many options have lots of overlaps in hyperparameter settings and architecture sources. These relationships can help us to obtain more effective embedding representations, help AutoTS to understand the composition characteristics of each module option more deeply, and thus make more effective decisions. Based on this idea, in AutoTS, we design an embedding learning method based on knowledge graph to learn higher-level option embeddings from these relationships. 

\begin{figure}
	\centering
	\includegraphics[width=0.5\textwidth]{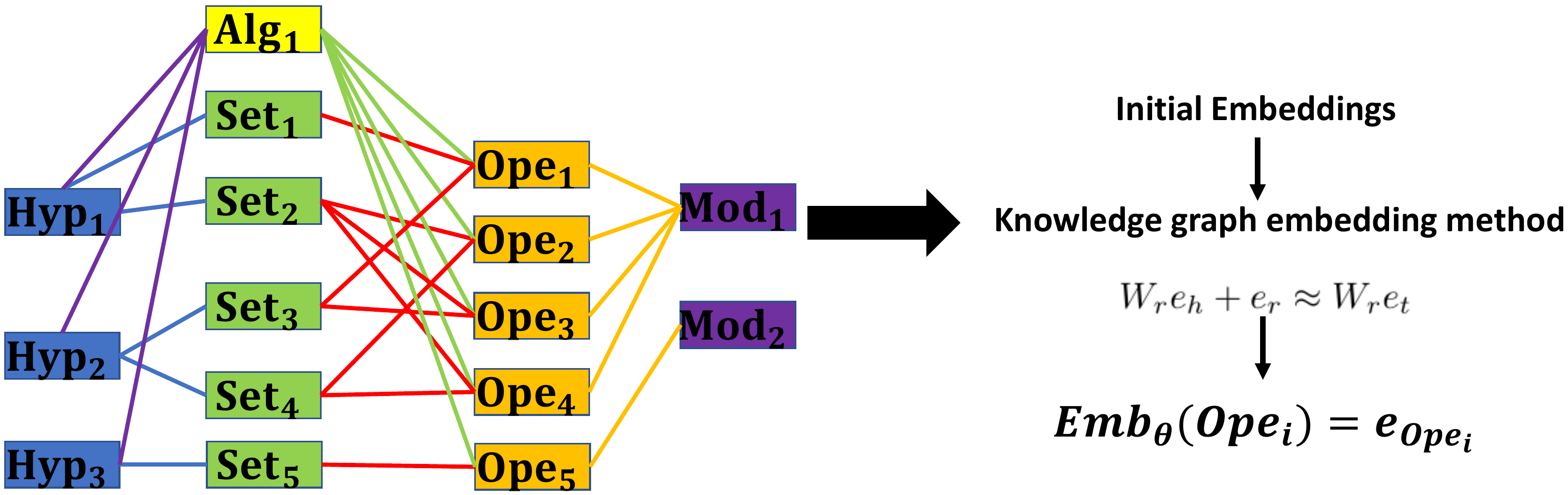}
	\caption{An example of the knowledge graph $\mathbb{G}$ in AutoTS. The rectangles with different colors represent the different types of entity nodes and the lines with different colors represent different types of entity relations.}
	\label{fig5}
\end{figure}

We aim to propose to construct a knowledge graph $\mathbb{G}$ to clarify the constituent relationship between various options and clarify their constituent differences and commonalities. Then, introduce a knowledge graph embedding learning method to add these constituent relationships into the embedding representation of each option and thus get a more effective $\mathbf{Emb_{\theta}}$.

\textbf{Knowledge Graph Construction.} The knowledge graph $\mathbb{G}$ contains five types of entity nodes: (1) TSF algorithm namely the architecture source ($\mathbf{Alg}$); (2) module name ($\mathbf{Mod}$); (3) hyperparameter name ($\mathbf{Hyp}$); (4) hyperparameter setting ($\mathbf{Set}$); (5) module option ($\mathbf{Opt}$). And it includes the following five types of entity relations: 
\begin{itemize}
\item $\mathbf{R_{1}}$: $\mathbf{Alg}\gets \mathbf{Opt}$, which indicates the architecture source of the module option;
\item $\mathbf{R_2}$: $\mathbf{Set}\gets \mathbf{Opt}$, which indicates the hyperparameter setting of the module option;
\item $\mathbf{R_3}$: $\mathbf{Mod}\gets \mathbf{Opt}$, which indicates the module that the module option belongs to;
\item $\mathbf{R_4}$: $\mathbf{Hyp}\gets \mathbf{Set}$, which denotes the name of the hyperparameter that the setting belongs to;
\item $\mathbf{R_5}$: $\mathbf{Alg}\gets \mathbf{Hyp}$, which denotes the TSF algorithm that the hyperparameter belongs to.
\end{itemize}
These relationships can help our algorithm better understand the composition characteristics of each module option: $\mathbf{R_1}\sim \mathbf{R_3}$ describe the composition details of module options; $\mathbf{R_5}\sim \mathbf{R_6}$ illustrate the meaning and source of each hyperparameter setting. Fig.~\ref{fig5} is an example of $\mathbb{G}$.

\textbf{Embedding Learning Based on Knowledge Graph.} In AutoTS, we use TransR~\cite{DBLP:conf/aaai/LinLSLZ15}, a widely used knowledge graph embedding method, to effectively parameterize entities and relations in $\mathbb{G}$ as vector representations. Specifically, given a triplet $(h, r, t)$ in $\mathbb{G}$, we learn the embedding of each entity and relation by optimizing the translation principle:
\begin{equation}
W_{r}e_{h}+e_{r}\approx W_{r}e_{t}
\end{equation}
where $e_h,e_t \in \mathbb{R}^d$ and $e_r \in \mathbb{R}^k$ are the embedding for $h, t$, and $r$ respectively; $W_r \in \mathbb{R}^{k\times d}$ is the transformation matrix of relation $r$. 

\begin{algorithm}[t]
	\caption{AutoTS Algorithm}
	\begin{algorithmic}[1] 
		\REQUIRE Search space $\mathcal{S}=\mathcal{O}_{1}\wedge\mathcal{O}_{2}\wedge\cdots\wedge\mathcal{O}_{7}$, 
		Search time $\mathcal{T}$, Number of top models $K$, Random possibility $\epsilon$, Prune ratio $\gamma$, Embedding size $N$
		
		\STATE $\mathbb{G}\gets$ Knowledge graph on $\sum_{i=1}^{7}|\mathcal{O}_{i}|$ operations in $\mathcal{S}$
		\STATE $\mathbf{Emb}_{\theta}\in \mathbb{R}^{N\times \sum_{i=1}^{7} |\mathcal{O}_{i}|}\gets$ Learn embeddings of  module options involved in $\mathcal{S}$ according to TransR and $\mathbb{G}$
		\STATE Execute Vertical Pruning Method in Algorithm~\ref{alg1}, get $\mathbf{BestMs}$, $\mathbf{Emb}_{\theta}$, and $\mathcal{D}$
		\STATE Execute Horizontal Pruning Method in Algorithm~\ref{alg2}, get $\mathbf{BestMs}$
		\STATE $\textbf{BestM}\gets$ Optimal model code in $\mathbf{BestMs}$
		\RETURN $\textbf{BestM}$
\end{algorithmic}  
\label{alg3}
\end{algorithm}

We extract the well-trained entity embeddings w.r.t. the module option to initialize the $\mathbf{Emb_{\theta}}$. Then, we apply $\mathbf{Emb_{\theta}}$ to the two-stage strategy, allowing $\mathbf{Emb_{\theta}}$ to be further optimized by learning performance characteristics of the module options (Equation~\ref{equ4}), so as to achieve better embedding representations of the module options.

\textbf{Advantage Analysis.} With the addition of the knowledge graph based embedding learning method, AutoTS can have a better understanding of the characteristics of various options in the search space, i.e., understanding module options from two aspects: performance and composition, so as to get a more effective $\mathbf{Emb_{\theta}}$. With this $\mathbf{Emb_{\theta}}$, AutoTS can make more effective decisions on the importance evaluation of options and the performance analysis of the model code, and thus achieve better search results.

\subsection{AutoTS Algorithm}\label{section:4.4}

Combining the methods in Section~\ref{section:4.2} and Section~\ref{section:4.3}, we get the AutoTS algorithm to quickly explore the TSF search space defined in Section~\ref{section:4.1} and realize the efficient and automatic design of TSF model. Algorithm~\ref{alg3} gives the complete pseudo code of AutoTS.

\section{Experiments}\label{section:5}
In this section, we examine the performance of AutoTS. We compare AutoTS with existing NAS algorithms and the state-of-the-art TSF models. In addition, ablation experiments are conducted to analyze the two-stage pruning method and the knowledge graph based embedding learning method in AutoTS. All experiments are implemented using Pytorch.

\subsection{Experimental Setup}\label{section:5.1}

\textbf{Datasets.} We conducted our experiments upon the single-step multivariate time series datasets. We compared several of models on these real-world datasets. Traffic dataset is from the California Department of Transportation which contains road occupancy rates measured by 862 sensors in San Francisco Bay area freeways during 2015 and 2016. Solar-Energy is collected by the National Renewable Energy Laboratory which is the solar power output from the 137 PV plants in Alabama State in 2007. Electricity contains the electricity sonsumption for 321 clients from 2012 to 2014 collected by the UCI Machine Learning Repository. Exchange-Rate consists of the daily exchange rates of eight foreign countries including Australia, British, Canada, Switzerland, China, Japan, New Zealand, and Singapore ranging from 1990 to 2016. We split these four datasets into a training set (60\%), validation set (20\%) and test set (20\%).

\begin{table*}[htbp]
	\centering
	\caption{Dataset statistics.}
	\begin{tabular}{ccccc}
		\toprule[1.5pt]
		Datasets & Samples&Input Length&Output Length&Variables \\
		\midrule[1pt]
		\#Solar-Energy&52,560&168&1&137\\
		\#Traffic&17544&168&1&862\\
		\#Electricity&26304&168&1&321\\
		\#Exchange-Rate&7588&168&1&8\\
		\bottomrule[1.5pt]
	\end{tabular}
\end{table*}

\textbf{Compared Methods.}  To clearly discover the improvment in the model architecture through the AutoML, we compare our AutoML model with the resource algorithm. They are also the SOTA algorithms in time series forecasting. Hence comparing with them will lead to a convincing results. We also compare our model with the widely used algorithms (RNN,TCN).
\begin{itemize}
	\item RNN: Convolutional nerual networks.
	\item ForecastNet: A time-variant module with deep feed-forward architecture.
	\item TPA-LSTM: A module with a set of filters to extract time-invariant temporal patterns which is similar to transforming time series data into its “frequency domain” and a novel attention mechanism
	\item TCN: Temporal convolutional networks.
	\item N-BEATS: A deep neural architecture based on backward and forward residual links and a very deep stack of fully-connected layers.
	\item MTGNN:.  A deep network with graph learning module and novel mix-hoppropagation layers and  dilated inception layers. And the graph learning, graph convolution, and temporal convolution modules are jointly learned in the end-to-end framework.
\end{itemize}
In addition, we compare AutoTS with two popular search strategies for AutoML: a RL search strategy that combines recurrent neural network controller and a commonly used baseline in AutoML, Random Search. We replece their search space to our search space, so as to examine their performance under the automatic TSF model deign tasks.

\textbf{Implementation Details.} When training the model, we apply Adam optimizer, learning rate $lr=0.001$ and $epochs = 10$.  We set horizon $\Delta=3,6,12,24$ to explore the capacity of the model for dealing with different forecasting horizon. 

\textbf{Evaluation Metrics.} We use two metrics to evaluate the performance including Root Relative Squared Error (RRSE) and Empirical Correlation Coefficient (CORR). For RRSE, the lower is the better. For CORR, the higher is the better. 

\begin{figure}
	\centering
	\includegraphics[width=0.5\textwidth]{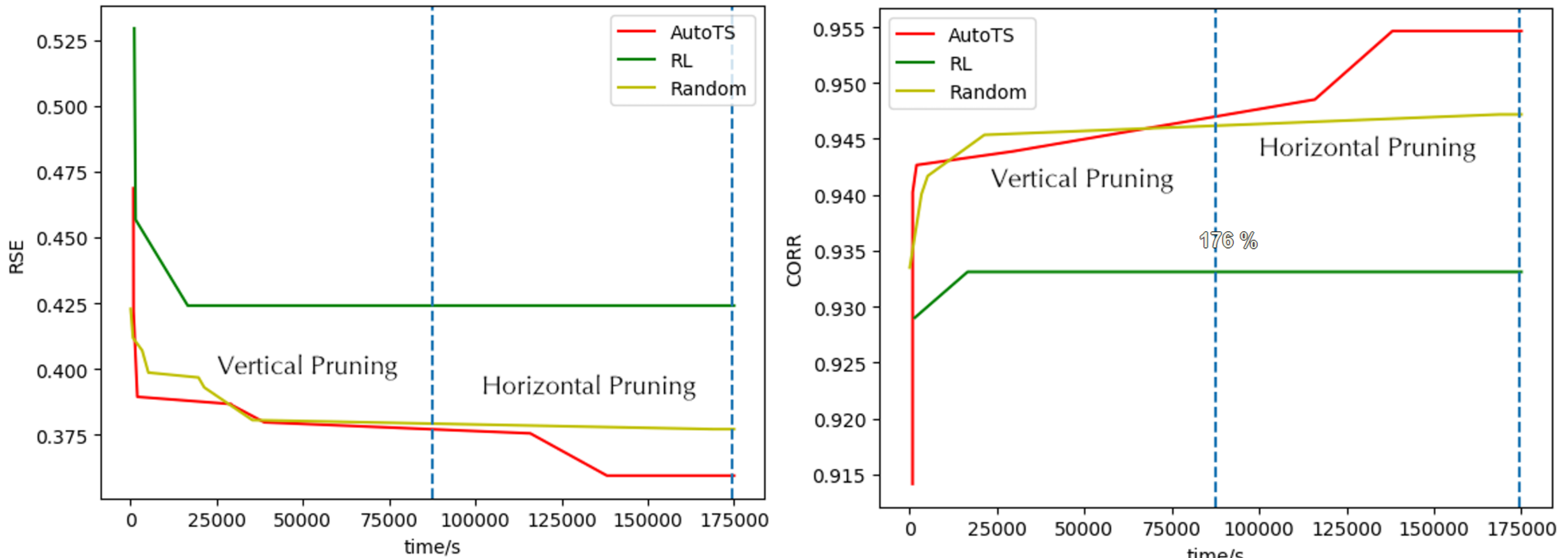}
	\caption{Comparison of different NAS algorithms on automated TSF model design task on Solar-Energy dataset.}
	\label{fig6}
\end{figure}

\subsection{Main Results}

We evaluate the performances of AutoTS, and compare it with the methods mentioned above. For each experiment, we use the same training param settings, in order to isolate the impact of the training loss and to enable fair comparisons. Overall results are presented in Table 3 and Fig.~\ref{fig6}.

Comparing with the source algorithms, the results show that AutoTS can help construct better model than them. It shows that we can get a better model by assembling the modules. And comparing with other searching strategies, two-stage pruning gives a remarkable improvement in the performance. We can clearly see the turning point of performance improvement from the search process diagram. In terms of results, AutoTS can explore the search space more deeply and get a better RSE and CORR value. In terms of time cost, from the process diagram we can see that the curve of AutoTS has more steep improving trend and has better performance than the other performance with the same searching time. 

In the diagram, we marked the boundary between vertical pruning and horizontal pruning. In the vertical pruning stage, it appears that the three algorithms have a normal trend of performance  improvement though the AutoTS has the faster speed. But in the horizontal pruning stage, with the extension of search time, the other two algorithms improve performance less and less, or even almost no longer. AutoTS manifests more promising. It still has a turning point in the horizontal stage and has a performance improvement. This indicates that our algorithm has stronger stamina and can explore search space more deeply.


\subsection{Ablation Study}

In this part, we further investigate the effect of transverse pruning vertical pruning strategy among each methods of AutoTS algorithm. So we use the following three variants of AutoTS to verify the innovations presented in this paper.

\begin{itemize}
\item[(1)] TransPruning: This algorithm deleted the transverse pruning while preserving the vertical pruning and the other AutoTS architecture.
\item[(2)] VertiPruning: This algorithm deleted the vertical pruning while preserving the transverse pruning and the other AutoTS architecture.
\item[(3)] NonePruning: This algorithm deleted the two-stage pruning while preserving the other AutoTS architecture.
\end{itemize}

Corresponding results are shown in Table 4. We can see that AutoTS performs much better than the TransPruning and the VertiPruning. This demonstrates us the importance and necessity of two-stage pruning strategy in our algorithm. As mentioned before, with the pruning strategy, we are able to explore modules as many as possible and at the same time find more flexible architecture efficiently. These two kinds of pruning optimize the model from two different aspects in AutoTS and show desirable results. Thus the combination of them is meaningful and necessity.

Moreover, we find that NonePruning performs much worse than AutoTS. We notice that it also decreases much the speed of searching. This result shows us that our pruning strategy is suitable
for our model search and helps usefully extract the characteristics of each module.

\section{Conclusion and Future Works}\label{section:6}

In this paper, we propose AutoTS algorithm to realize the automatic and efficient design of the TSF model. In AutoTS, we use the modular reorganization method to construct an effective search space suitable for the TSF area. In addition, we design the two-stage purning strategy and a knowledge graph analysis method to quickly understand the characteristics of search space components and prune the search space, so as to realize the efficient search of well-performed TSF models. The experimental results show that our proposed algorithm can quickly determine a high-quality and small-scale search space and design a better TSF model for a given data set in a shorter time. Compared with the existing NAS algorithm and manually designed TSF model, our proposed AutoTS is more effective and practical. It can greatly reduce the development cost of TSF model, so that ordinary users can easily use TSF models. In future work, we will try to consider adding more efficient TSF models to build our search space, and design more efficient search strategies to effectively deal with larger and richer search space.


%



\ifCLASSOPTIONcaptionsoff
  \newpage
\fi



%

%

\begin{IEEEbiography}[{\includegraphics[width=1in,height=1.25in,clip,keepaspectratio]{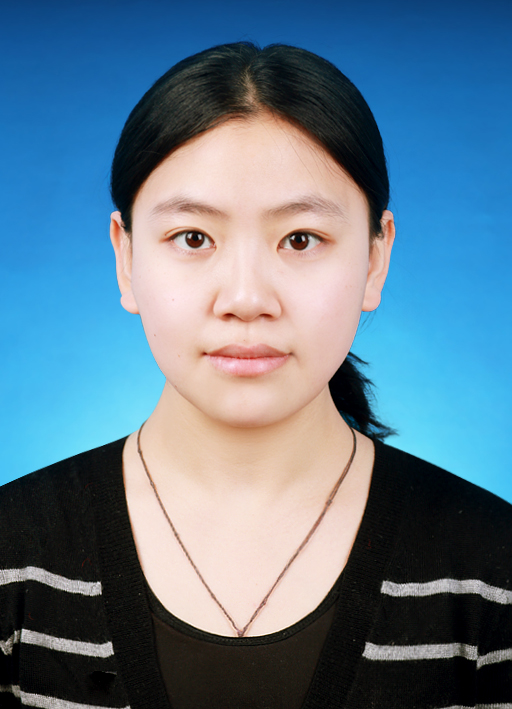}}]{Chunnan Wang}
received the BS degree in computer science from the Harbin Institute of Technology, in 2017. She is currently working toward the PhD degree in the School of Computer Science and Technology, the Harbin Institute of Technology, China. Her research interests include automated machine learning, graph data mining and federated learning.
\end{IEEEbiography}

\begin{IEEEbiography}[{\includegraphics[width=1in,height=1.25in,clip,keepaspectratio]{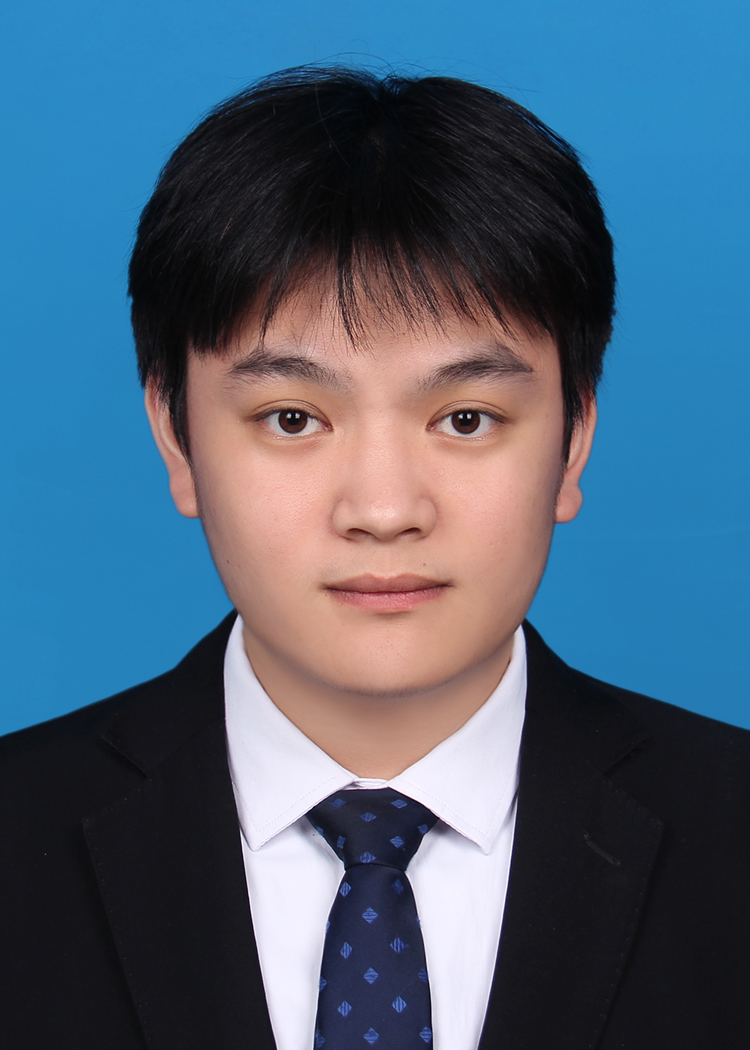}}]{Xingyu Chen}
is an undergraduate in the school of mechatronics engineering of Harbin Institute of Technology, China, in 2019. He has won the National Scholarship, four first class People's Scholarships, SMC Scholarship, and many outstanding awards of science and innovation competition. He is engaged in the research of time series prediction in the Massive Data Computing Research Center of Harbin Institute of Technology.
\end{IEEEbiography}

\begin{IEEEbiography}[{\includegraphics[width=1in,height=1.25in,clip,keepaspectratio]{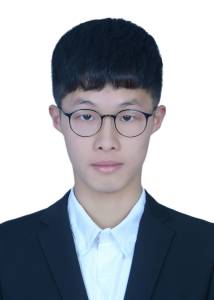}}]{Chengyue Wu}
is an undergraduate in the school of computer science of Harbin Institute of Technology, China, in 2019. He has won the National Scholarship, four first class People's Scholarships, Samsung Scholarship, and many outstanding awards of science and innovation competition. His research interests include machine learning and data mining. He has submitted a patent about AutoML for approval.
\end{IEEEbiography}

\begin{IEEEbiography}[{\includegraphics[width=1in,height=1.25in,clip,keepaspectratio]{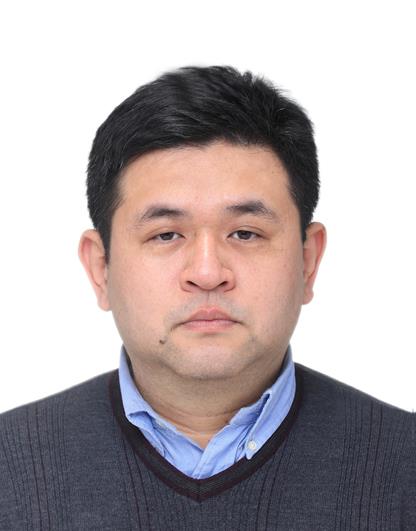}}]{Hongzhi Wang}
is a professor of department of Computer Science and Technology, Harbin Institute of Technology. He is the head of massive data computing center and the vice dean of the honors school of Harbin Institute of Technology, the secretary general of ACM SIGMOD China, outstanding CCF member, a standing committee member CCF databases and a member of CCF big data committee. His research fields include database management system, big data quality, big data analysis and mining, AI for DB, DB for AI, Industrial Big Data, Blockchain and Financial Big Data and automated machine learning.
\end{IEEEbiography}




\end{document}